\documentclass[runningheads]{llncs}

\usepackage{eccv}

\usepackage{eccvabbrv}

\usepackage{graphicx}
\usepackage{booktabs}
\usepackage{multirow}
\usepackage[table]{xcolor}

\usepackage[accsupp]{axessibility}  

\newcommand{\todo}[1]{\textcolor{red}}
\newcommand{\tong}[1]{\textcolor{violet}}
\newcommand{\patsorn}[1]{\textcolor{pink}}

\newcommand{\warning}[1]{{\it\color{red}}}

\renewcommand{\warning}[1]{}
\renewcommand{\todo}[1]{}
\renewcommand{\patsorn}[1]{}
\renewcommand{\tong}[1]{}

\usepackage{hyperref}

\usepackage{orcidlink}

\begin{document}

\title{Infinite Gaze Generation for Videos with Autoregressive Diffusion} 
%
%
\author{First Author\inst{1}\orcidlink{0000-1111-2222-3333} \and
Second Author\inst{2,3}\orcidlink{1111-2222-3333-4444} \and
Third Author\inst{3}\orcidlink{2222--3333-4444-5555}}
%
%
\author{Jenna Kang\inst{1} \and
Colin Groth\inst{1} \and
Tong Wu\inst{2} \and
Finley Torrens\inst{1} \and
Patsorn Sangkloy\inst{1} \and
Gordon Wetzstein\inst{2} \and
Qi Sun\inst{1}}

\institute{
New York University, New York, NY, USA\\
\email{jennakang@nyu.edu, c.groth@nyu.edu, ft2377@nyu.edu, patsorn.sangkloy@gmail.com, qisun@nyu.edu}
\and
Stanford University, Stanford, CA, USA\\
\email{wutong16.thu@gmail.com, gordonwz@stanford.edu}
}

\maketitle
\begin{abstract}
%
Predicting human gaze in video is fundamental to advancing scene understanding and multimodal interaction. While traditional saliency maps provide spatial probability distributions and scanpaths offer ordered fixations, both abstractions often collapse the fine-grained temporal dynamics of raw gaze. Furthermore, existing models are typically constrained to short-term windows ($\approx$ 3--5s), failing to capture the long-range behavioral dependencies inherent in real-world content. We present a generative framework for infinite-horizon raw gaze prediction in videos of arbitrary length. By leveraging an autoregressive diffusion model, we synthesize gaze trajectories characterized by continuous spatial coordinates and high-resolution timestamps. Our model is conditioned on a saliency-aware visual latent space. Quantitative and qualitative evaluations demonstrate that our approach significantly outperforms existing approaches in long-range spatio-temporal accuracy and trajectory realism. Project website: 
\url{https://www.immersivecomputinglab.org/publication/infinite-gaze/}.
%


%
%
%
%
\keywords{gaze prediction, videos, saliency, scanpath, diffusion models, autoregression}
\end{abstract}
\section{Introduction}
\label{sec:intro}

Modeling human visual attention through spatial saliency and temporal scanpaths is vital for scene understanding, enabling applications from foveated rendering \cite{zeng2025foveated,patney2016towards} to intention-aware robotics \cite{li2022fusion,zhang2023human}. However, the systematic collection of high-fidelity gaze data is hindered by expensive hardware requirements, cumbersome calibration, and privacy constraints. This data scarcity necessitates generative methods capable of synthesizing plausible, realistic gaze behavior. By leveraging a latent world model to understand evolving environmental dynamics, such approaches can move beyond static maps to produce continuous, infinite-horizon trajectories that accurately reflect human visual engagement with the world.

Existing gaze prediction models generally fall into two categories: static saliency maps, which provide time-independent spatial probability distributions \cite{jiang2018deepvs,wang2019revisiting}, and scanpaths, which discretize eye movements into sequences of fixations \cite{karadiffeye,martin2022scangan360}. While these abstractions are effective for images, they often collapse the high-frequency temporal dynamics of raw eye-tracking data and struggle to represent smooth pursuit in video. Furthermore, current models are typically constrained to short-term viewing windows ($\approx$ 3--5s), failing to capture the long-range behavioral dependencies found in real-world, long-form content. These limitations prevent existing methods from synthesizing the continuous, infinite-horizon gaze trajectories required for deep scene understanding and realistic dynamic interaction.


Inspired by recent advances in video world models that generate content in a causal manner \cite{po2025long,wu2025video,ding2025understanding}, we process video sequences frame-by-frame to synthesize gaze data autoregressively. Our approach predicts gaze for the current frames by conditioning on visual context, as well as the history of previously generated gaze coordinates. Notably, recognizing that visual saliency is a stronger prior than general visual features for inducing eye movements, we introduce the idea of applying saliency-aware latents to encode visual frame information. This causal formulation ensures temporal continuity and allows the model to act as a latent world model, capable of generating realistic, infinite-horizon gaze trajectories that remain grounded in the evolving scene dynamics.

Through extensive quantitative evaluations, qualitative analysis, and human subject studies, we demonstrate that our approach significantly advances the state-of-the-art in generating accurate and human-like spatio-temporal gaze trajectories for videos of arbitrary length. Our results confirm that the model effectively maintains long-range consistency while capturing the nuanced, high-frequency dynamics characteristic of real human vision. We envision this research establishing a new frontier in synthesizing natural human behaviors by leveraging the power of emerging autoregressive world models. To this aim, we will release our code and models upon acceptance. 

\section{Related Work}

\subsection{Scanpath Generation for Images}

Visual attention acts as the selective mechanism that determines which parts of a scene humans prioritize \cite{kruijne2016implicit}. Early computational models focused on predicting spatial saliency, estimating which regions of an image are most likely to attract human gaze \cite{Judd_2009, bylinskii2016where, borji2021saliency, borji2013state}. However, saliency maps ignore the temporal structure of eye movements and do not capture the order or timing of fixations, which are fundamental aspects of human visual behavior.

To address this limitation, researchers formulated the problem as scanpath prediction, modeling eye movements as sequences of discrete fixations \cite{martin2022scangan360, kummerer2022deepgaze}. Early deep learning approaches used Recurrent Neural Networks (RNNs) to capture dependencies between successive fixations \cite{chen2021predicting, zelinsky2019benchmarking, Assens2018pathgan, Sun2019VisualSP}. More recently, Transformer-based architectures have achieved state-of-the-art performance by better modeling long-range spatial and temporal dependencies in scanpath sequences \cite{fang2024oat, nishiyasu2024gaze, yang2024unifying, xue2025few}.

Despite these advances, representing gaze as discrete fixation sequences simplifies the rich dynamics present in raw eye-tracking data. To better capture fine-grained eye movement behavior, recent work has explored continuous gaze trajectory generation. For example, DiffEye \cite{karadiffeye} formulates gaze prediction as a diffusion process that generates fine-grained gaze trajectories conditioned on patch-level visual features, grounding the predicted gaze coordinates in the semantic structure of natural images.

\subsection{Scanpath Modeling in 360-Degree Content}

Modeling gaze behavior in immersive environments introduces additional challenges due to spherical geometry and the wider field of view. Several works extend scanpath prediction methods to 360-degree images, where gaze trajectories are represented as fixation sequences on a spherical panorama. For example, ScanDMM \cite{scanddm} is a probabilistic deep Markov model that generates realistic variable-length 3D scanpaths on panoramic images by modeling time-dependent attention dynamics and visual working memory. ScanGAN360 \cite{scangan360} uses a conditional generative adversarial network to synthesize human-like 3D scanpaths on 360-degree images using adversarial training with a DTW-based loss. Similarly, PathGAN \cite{Assens2018pathgan} generates plausible fixation sequences by learning the joint distribution of fixation locations and their temporal order conditioned on an input image and its saliency map. More recently, diffusion-based approaches have also been applied to immersive content. DiffGaze \cite{jiao25diffgaze} generates gaze trajectories on 360-degree images using a diffusion model conditioned on a global spherical feature representation obtained through spherical convolution. 

In the case of 360-degree videos, many works focus on predicting viewer head motion or viewport trajectories rather than true eye movements. These methods treat gaze as spatiotemporal points on the spherical panorama corresponding to the user’s viewport or head orientation. For example, Xu et al. \cite{gaze360dataset} predict future gaze directions in 360-degree video sequences, while other works predict saliency or viewing patterns to optimize adaptive video streaming \cite{salientstreamnguyen,attention360video}. Transformer-based models such as VPT360 \cite{VTP360video} forecast future viewport trajectories using past viewing history, while diffusion-based approaches such as ScanTD \cite{scantd} generate plausible future scanpaths by denoising sequences of spherical fixation points. Other methods formulate panoramic video scanpath prediction as an information-theoretic problem \cite{scanpathcodelength} or incorporate multimodal signals such as motion \cite{scanpathmultimodalfusion}.

However, these 360-degree video approaches all rely on viewport or head-orientation signals rather than high-frequency eye-tracking trajectories. As a result, they are not directly comparable to datasets that record fine-grained 2D eye movements in conventional RGB videos.

\subsection{Autoregressive Diffusion and World Models}
Full-sequence diffusion models have demonstrated remarkable success in generating complex and high-dimensional data such as images \cite{ho2020denoising, song2021scorebased, rombach2021highresolution, Peebles2022DiT}, videos \cite{ho2022video, singer2022makeavideo, openai2024sora}, and continuous planning \cite{janner2022diffuser, dong2024diffuserlite}. However, they are typically constrained to fixed-length windows and struggle with variable-length generation. To unify the sequential flexibility of autoregressive next-token prediction with continuous diffusion, Diffusion Forcing \cite{chen2025diffusion} introduces a training paradigm that trains causal models to denoise future tokens while conditioning on a clean or partially noised past, stabilizing continuous long-horizon rollouts. This autoregressive diffusion paradigm has been particularly impactful in the development of generative world models, which require continuous, long-term predictions of physical environments and spatial trajectories. However, while these emerging models focus on generating long and variable-length videos \cite{yin2025causvid}, consistent environments \cite{wu2025video}, or autonomous agent actions \cite{zhang2025epona}, our work redirects this autoregressive diffusion paradigm toward modeling human eye movement reactions to the dynamic world.
\section{Method}
\begin{figure}[t]
    \centering
    \begin{subfigure}{0.56\linewidth}
        \centering
        \includegraphics[width=\linewidth]{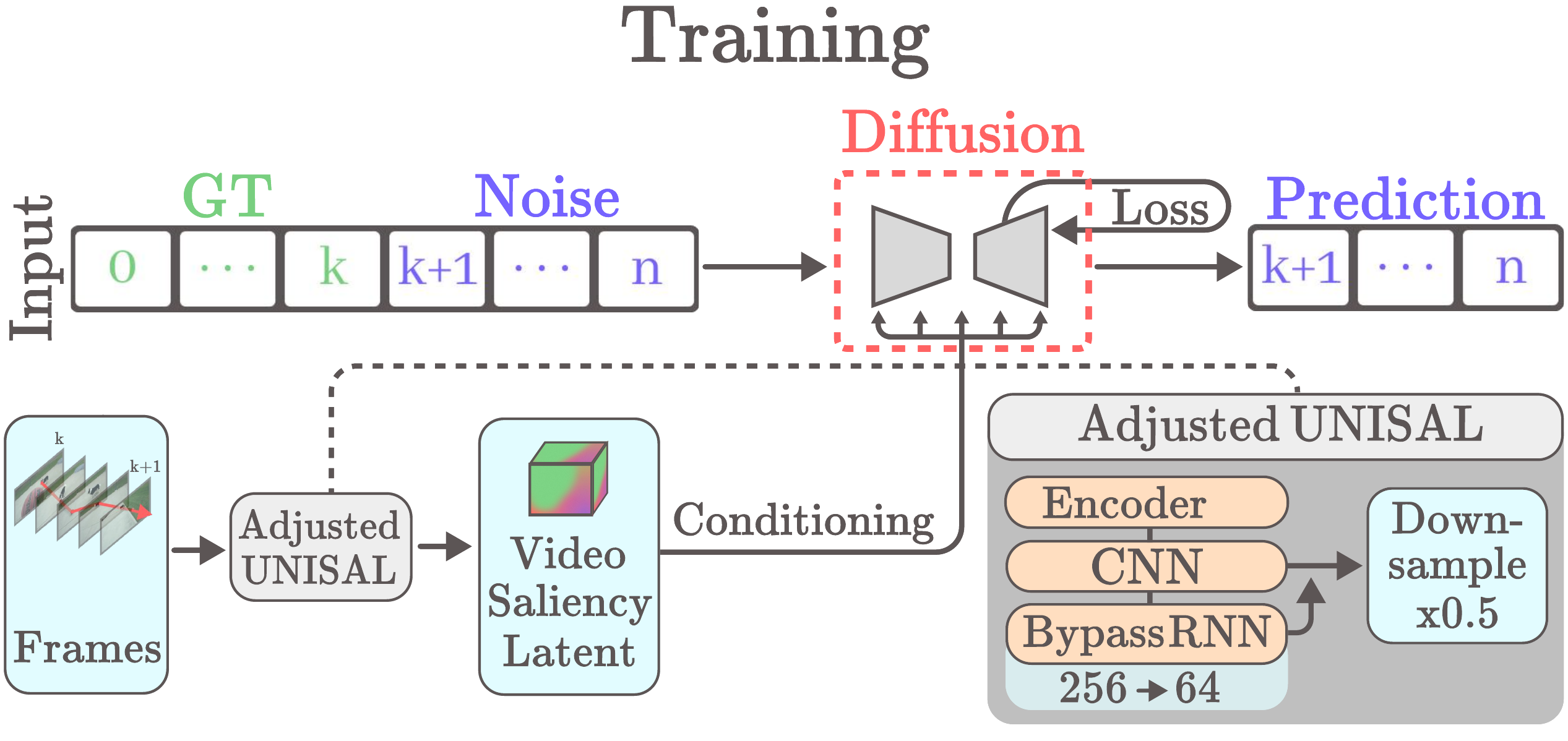}
        \caption{Training process.}
        \label{fig:model:training}
    \end{subfigure}
    \hfill
    \begin{subfigure}{0.43\linewidth}
        \centering
        \includegraphics[width=\linewidth]{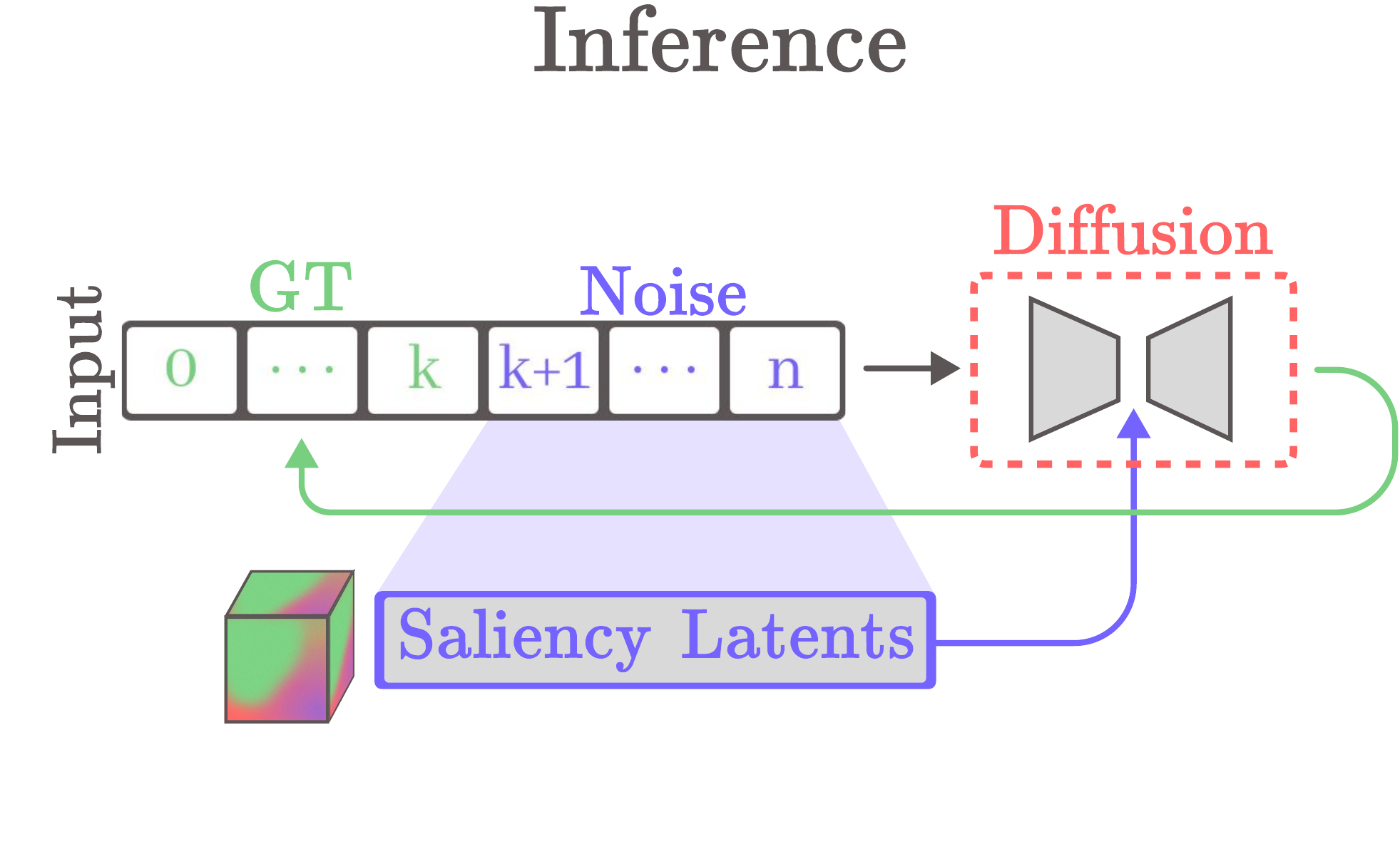}
        \caption{Inference process.}
        \label{fig:model:inference}
    \end{subfigure}
    \caption{\textit{Our training and inference pipelines.} The input of each iteration consists of the n-dimensional input vector with gaze history (GT) and noise for future prediction, and video frames corresponding to the prediction window. During inference, the gaze history of this input vector is filled with former predictions in an autoregressive manner. Video latents are generated frame-by-frame with a modified version of UNISAL. The blue boxes indicate our modifications, while the orange boxes indicate the original architecture. The low-dimensional video latents are processed by the U-Net with cross-attention at each block.}
    \label{fig:model}
\end{figure}

\subsection{Problem Formulation}
Let a video stimulus be represented as a sequence of RGB frames $V = \{ I_t \}_{t=1}^{T},$ $\quad I_t \in \mathbb{R}^{H \times W \times 3}$, where \(T\) denotes the number of frames, our goal is to perform gaze prediction $R \in \mathbb{R}^{T\times2}$. The gaze prediction at each timestep is a two-dimensional coordinate $x, y$, representing the spatial location in the frame.
Following DiffEye\cite{karadiffeye}, we model gaze prediction as learning a conditional generative model
$p_\theta(R \mid V)$ using a denoising diffusion probabilistic model (DDPM), which approximates the conditional distribution of gaze trajectories.

We further extend the gaze generator to operate in an autoregressive manner by introducing a historical gaze cache, with a subsequence of previously observed gaze coordinates. The historical sequence is concatenated before the predicted gaze trajectory to encourage temporal continuity and smooth transitions.

\subsection{Model Architecture}
As shown in Figure~\ref{fig:model}, we follow DiffEye \cite{karadiffeye} and adopt the same U-Net-based diffusion backbone 
as the noise prediction network. The model consists of hierarchical 1D convolutional blocks with multi-scale self-attention, and is conditioned on the diffusion timestep via sinusoidal embeddings. 
To extend the framework from static images to potentially infinite video sequences, we introduce two key modifications.
First, we reformulate gaze generation in an \textbf{autoregressive manner}, allowing the model to iteratively predict future gaze trajectories conditioned on previously generated gaze coordinates.
Second, due to the significantly larger information content in videos compared to images, we design an effective \textbf{stimulus conditioning} mechanism to make use of saliency latents and compress and aggregate temporal visual information before injecting it into the diffusion backbone.

\paragraph{\textbf{Autoregressive Generation}}
We adopt an autoregressive formulation similar to prior video diffusion models. 
Given a video sequence of length $n$, the first $k$ gaze coordinates serve as conditional prefix, while the remaining $n-k$ coordinates are generated.
During training, the windows of length $k$ are randomly sampled from the ground-truth segments. The model conditions on the historical gaze prefix 
$R_{1:k}$ and predicts the future gaze sequence $R_{k+1:n}$. The diffusion loss is computed only on the predicted segment.
At inference time, the model iteratively uses the most recent $k$ generated gaze coordinates as conditioning input to predict subsequent gaze points, thereby enabling long-horizon autoregressive gaze generation. Following prior measurements of visual working-memory loss over 1--6 s\cite{ricker2010loss}, we chose $k = 90$-frames (3 s) history as a midpoint balancing temporal context, noise, and compute cost.

\paragraph{\textbf{Stimuli Conditioning}}
To handle continuous video streams, an effective conditioning mechanism is crucial. Compared to static images, videos contain significantly larger information volume, making direct RGB conditioning inefficient and computationally expensive. We therefore emphasize information compression as a key design principle.

Instead of conditioning on raw RGB frames, we leverage video saliency maps as stimulus representation. 
Saliency provides a compact, single-channel description of the probability that an average viewer attends to a given spatial region. This representation is both semantically meaningful and content-independent, since saliency values directly correspond to gaze likelihood rather than raw pixel appearance.
We adopt UNISAL \cite{UNISAL}, a well-established video saliency model, as the backbone for saliency extraction. 
However, existing saliency models are designed for prediction accuracy rather than compact representation. To better align with our conditioning objective, we introduce a compressed bottleneck within the encoder–decoder architecture, as shown in Figure~\ref{fig:model}.
Specifically, the RNN channels are decreased from 256 to 64, which still provides enough temporal information yet saves significant size.
The latent after the encoder is further compressed in its spatial dimensions through 2D average pooling.
The compressed saliency latent is used as conditioning input to the diffusion backbone. 
During training and inference, saliency features are extracted per frame and concatenated temporally, 
while conditioning is applied at a reduced temporal frequency (every 5 frames) to further limit computational overhead. 
This design enables efficient long-horizon gaze generation without sacrificing performance.

\section{Experiments}

\subsection{Experimental Setup}
\begin{figure}[t]
    \centering
    \includegraphics[width=\textwidth]{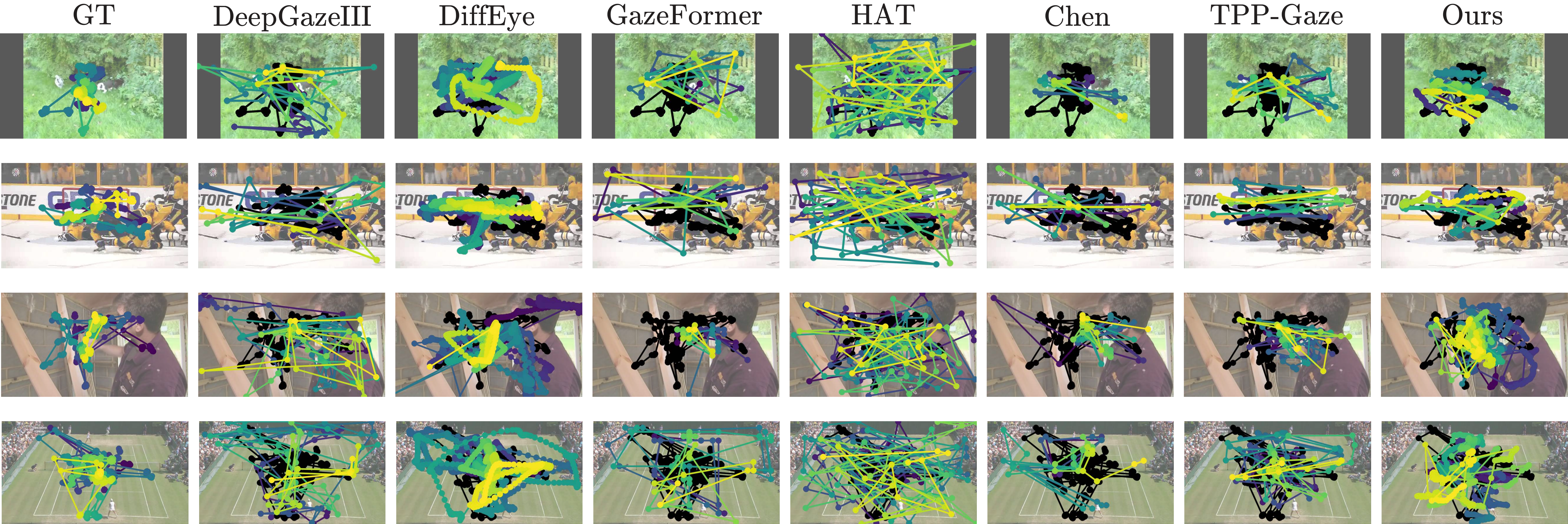}
    \caption{\textit{Comparison of the trajectories of video gaze predictions with different methods.} The image shows a representative frame selected from the video, but the path is generated for the entire duration of the video. Generated trajectories (color map indicates timestamps) are overlaid on the ground truth human gaze (black trajectories, which are also shown in the left column with the color map).}
    \label{fig:gaze_full}
\end{figure}

\paragraph{\textbf{Implementation Details}}
The model was trained for 70 epochs using the Adam optimizer with a constant learning rate of $1 \times 10^{-4}$. We employed a DDPM linear noise schedule, spanning from $1 \times 10^{-4}$ to $2 \times 10^{-2}$. 
For sampling, we adopted Denoising Diffusion Implicit Models (DDIM) \cite{song2020denoising} to achieve high-quality generation in significantly fewer steps without compromising performance. Accordingly, the number of diffusion steps was set to 1000 during training and reduced to 50 during sampling for improved efficiency.

\paragraph{\textbf{Datasets}}
We evaluate the proposed method on two public video datasets that provide diverse visual content and human gaze annotations.
We use \textit{DIEM} as the in-domain benchmark \cite{mital2011clustering}. The dataset contains 84 videos with a resolution of 1280×720 and durations between 30 to 209 seconds. Each video provides gaze data from 67 observers on average, with a total viewing time of 553,116 seconds (16,593,480 frames). In comparison, the only other gaze prediction method that outputs raw gaze trajectories (DiffEye \cite{karadiffeye}) is trained on 45,135 seconds of viewing time (MIT1003 Dataset \cite{mit1003}). We adopt an 80/20 train–test split. We select DIEM for training due to its comparatively high spatial resolution, long temporal extent, and dense gaze annotations.

We further evaluate generalization on \textit{DHF1K} \cite{wang2018revisiting}. The dataset comprises 1,000 videos with durations between 17 and 42 seconds and a spatial resolution of 640×360. Each video includes gaze data from 17 observers. The content spans a wide range of scene motion and camera motion patterns.
For evaluation, we select 16 videos, with four videos from each motion category defined by the authors: slow camera motion, fast camera motion, slow scene motion, and fast scene motion. We manually verify that the randomly selected subset covers diverse content and motion characteristics. 

\paragraph{\textbf{Evaluation Metrics}}

We evaluate the results using four standard metrics from path assessment of video data \cite{scanpathcodelength}.
(1) \textit{Levenshtein Distance} measures the sequence similarity based on the number of insertion, deletion, and substitution operations required to match a GT \cite{privitera2002algorithms};
(2) \textit{Discrete Fréchet Distance} measures the similarity between two curves by evaluating the minimal maximum pointwise distance under a monotonic traversal of both sequences \cite{aronov2006frechet};
(3) \textit{Dynamic Time Warping} (DTW) computes an optimal alignment between two sequences that may differ in speed or length by minimizing cumulative alignment cost \cite{berndt1994using};
(4) \textit{Maximum Temporal Correlation} quantifies the highest cross-correlation between two temporal signals over possible time shifts \cite{scanpathcodelength}.

Our evaluation protocol follows established procedures in prior work on image-based scanpath prediction \cite{scantd, jiao25diffgaze}.
For each video, we generate 10 gaze trajectories per model. Given the multiple ground-truth (GT) gaze paths per video (see Dataset section for numbers), we compute the evaluation metrics by performing pairwise comparisons between each GT path and all 10 predicted trajectories.
Metric scores are first computed per GT path. We report two variants. The best score selects the maximum (or minimum, depending on the metric direction) value among the 10 pairwise comparisons. The mean score corresponds to the average over all 10 comparisons. The best score reflects the peak performance of a method under multiple samples. The mean score quantifies the consistency of the predicted trajectories.
Finally, for each metric, we average both the best and mean scores across all GT gaze paths and all videos in the dataset.

\paragraph{\textbf{Comparison Methods}}
\begin{figure}
    \centering
    \includegraphics[height=.9\textheight]{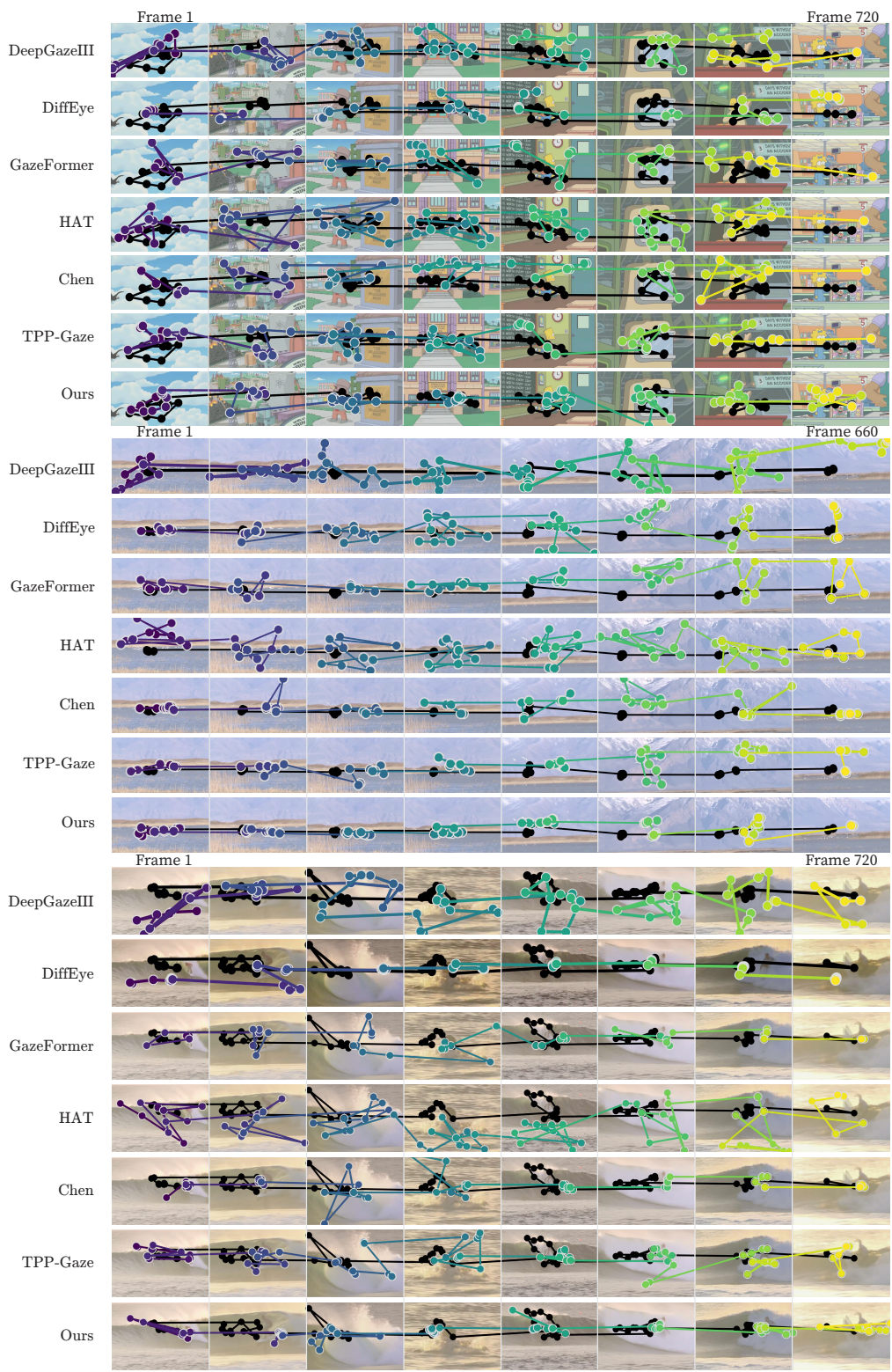}
    \caption{\textit{Temporal progression of video gaze predictions with different methods.} Generated trajectories are overlaid on the ground-truth human gaze (black). Note the closer proximity between our method and the ground truth, indicating greater similarity compared with alternative approaches.}
    \label{fig:gaze_example}
\end{figure}
Since our work introduces the first video-based gaze prediction model, no direct baselines exist. We adapt existing state-of-the-art image-based scanpath and gaze prediction models to the video domain for comparison.
The baselines include the transformer-based architectures Human Attention Transformer (\textit{HAT}) \cite{yang2024unifying} and \textit{GazeFormer} \cite{gazeformer}, \textit{TPP-Gaze} which is based on neural temporal point processes \cite{d2025tpp}, the convolutional network \textit{DeepGaze III} \cite{deepgaze3}, the reinforcement learning model of Chen et al. (originally designed for visual question answering) \cite{chen2021predicting}, and the diffusion-based method \textit{DiffEye} \cite{karadiffeye}. Among these, DiffEye is the only method that predicts dense, frame-to-frame gaze motion.
To ensure fair comparison, we post-process the sample density of all scanpath prediction models to match the number of samples from our method (30 samples per second) by duplicating fixation positions proportionally. Since DiffEye produces raw gaze at a higher sampling rate than our method, we evenly downsample its gaze sequence. All comparison method predictions are generated with their default 3-second length. However, videos are typically much longer.
To achieve autoregressive behavior for longer sequences, we process the video in 3-second chunks and concatenate outputs. For methods allowing us to control the gaze starting position (HAT and DeepGaze III), we use the last predicted gaze of the previous chunk as the initial position for the next chunk. For each 3-second segment, the first frame is provided as input, since the baseline models are single-image based. For multi-task models, we ensure that scanpaths are generated for the free-viewing task specifically. For additional fairness, we also evaluate all methods within their native 3-second window without autoregressive adjustment. Additionally, we retrain DiffEye using the same saliency latents and no autoregression to demonstrate that saliency conditioning alone is insufficient for long-horizon video gaze generation.

\definecolor{rowgray}{gray}{0.92}
\definecolor{rowgray2}{gray}{0.86}
\definecolor{bestcell}{gray}{0.75}

\begin{table*}[t]
\centering
\small
\caption{\textit{Scanpath evaluation on DIEM and DHF1K test sets.} Lower is better for all metrics except Maximum Temporal Correlation. Metrics are scaled as indicated in the column headers. \textbf{Bold} values indicate the best scores, while \underline{underlined} values denote the second-best scores.}
\resizebox{\textwidth}{!}{
\setlength{\tabcolsep}{12pt}
\begin{tabular}{l l cccccccc}
\toprule
\multirow{2}{*}{Dataset} & \multirow{2}{*}{Method} 
& \multicolumn{2}{c}{Levenshtein $\downarrow$}
& \multicolumn{2}{c}{Disc. Fréchet ($\times 10^2$) $\downarrow$}
& \multicolumn{2}{c}{DTW ($\times 10^4$) $\downarrow$}
& \multicolumn{2}{c}{Max. Temp. Corr. $\uparrow$} \\

& 
& Mean & Best 
& Mean & Best
& Mean & Best
& Mean & Best \\
\midrule
\multirow{7}{*}{\textbf{DIEM}} & DeepGaze III \cite{deepgaze3}
& \cellcolor{rowgray}733 & \cellcolor{rowgray}711
& \cellcolor{rowgray}4.67 & \cellcolor{rowgray}3.96
& \cellcolor{rowgray}19.2 & \cellcolor{rowgray}16.9
& \cellcolor{rowgray}0.0954 & \cellcolor{rowgray}\underline{0.198} \\
& DiffEye \cite{karadiffeye}
& \cellcolor{rowgray2}787 & \cellcolor{rowgray2}776
& \cellcolor{rowgray2}4.33 & \cellcolor{rowgray2}3.73
& \cellcolor{bestcell}\textbf{14.4} & \cellcolor{rowgray2}13.2
& \cellcolor{rowgray2}\underline{0.106} & \cellcolor{rowgray2}0.173 \\
& GazeFormer \cite{gazeformer}
& \cellcolor{rowgray}\underline{704} & \cellcolor{rowgray}679
& \cellcolor{rowgray}4.38 & \cellcolor{rowgray}3.63
& \cellcolor{rowgray}15.7 & \cellcolor{rowgray}13.4
& \cellcolor{rowgray}0.0674 & \cellcolor{rowgray}0.143 \\
& HAT \cite{yang2024unifying}
& \cellcolor{rowgray2}748 & \cellcolor{rowgray2}731
& \cellcolor{rowgray2}4.95 & \cellcolor{rowgray2}4.29
& \cellcolor{rowgray2}20.8 & \cellcolor{rowgray2}18.9
& \cellcolor{rowgray2}0.0887 & \cellcolor{rowgray2}0.181 \\
& Chen et al. \cite{chen2021predicting}
& \cellcolor{rowgray}\underline{704} & \cellcolor{rowgray}\underline{676}
& \cellcolor{rowgray}4.36 & \cellcolor{rowgray}3.54
& \cellcolor{rowgray}15.5 & \cellcolor{rowgray}13.0
& \cellcolor{rowgray}0.0800 & \cellcolor{rowgray}0.187 \\
& TPP-Gaze \cite{d2025tpp}
& \cellcolor{rowgray2}713 & \cellcolor{rowgray2}678
& \cellcolor{bestcell}\textbf{3.81} & \cellcolor{bestcell}\textbf{3.10}
& \cellcolor{rowgray2}\underline{14.5} & \cellcolor{bestcell}\textbf{12.3}
& \cellcolor{rowgray2}0.0942 & \cellcolor{rowgray2}0.186 \\
& Ours
& \cellcolor{bestcell}\textbf{701} & \cellcolor{bestcell}\textbf{673}
& \cellcolor{rowgray}\underline{4.05} & \cellcolor{rowgray}\underline{3.28}
& \cellcolor{bestcell}\textbf{14.4} & \cellcolor{rowgray}\underline{12.5}
& \cellcolor{bestcell}\textbf{0.197} & \cellcolor{bestcell}\textbf{0.286} \\
\midrule

\multirow{7}{*}{\textbf{DHF1K}}
& DeepGaze III \cite{deepgaze3}
& \cellcolor{rowgray}248 & \cellcolor{rowgray}237
& \cellcolor{rowgray}3.12 & \cellcolor{rowgray}2.67
& \cellcolor{rowgray}11.12 & \cellcolor{rowgray}9.41
& \cellcolor{rowgray}0.136 & \cellcolor{rowgray}0.317 \\

& DiffEye \cite{karadiffeye}
& \cellcolor{bestcell}\textbf{239} & \cellcolor{rowgray2}231
& \cellcolor{bestcell}\textbf{2.34} & \cellcolor{rowgray2}1.95
& \cellcolor{rowgray2}\underline{4.24} & \cellcolor{rowgray2}3.79
& \cellcolor{rowgray2}0.232 & \cellcolor{rowgray2}0.352 \\

& GazeFormer \cite{gazeformer}
& \cellcolor{rowgray}\underline{240} & \cellcolor{rowgray}229
& \cellcolor{rowgray}2.49 & \cellcolor{rowgray}1.93
& \cellcolor{rowgray}4.43 & \cellcolor{rowgray}3.69
& \cellcolor{rowgray}0.258 & \cellcolor{rowgray}0.392 \\

& HAT \cite{yang2024unifying}
& \cellcolor{rowgray2}250 & \cellcolor{rowgray2}243
& \cellcolor{rowgray2}3.12 & \cellcolor{rowgray2}2.67
& \cellcolor{rowgray2}6.08 & \cellcolor{rowgray2}5.43
& \cellcolor{rowgray2}0.179 & \cellcolor{rowgray2}0.289 \\

& Chen et al. \cite{chen2021predicting}
& \cellcolor{rowgray}241 & \cellcolor{rowgray}\underline{228}
& \cellcolor{rowgray}2.54 & \cellcolor{rowgray}\underline{1.84}
& \cellcolor{rowgray}4.70 & \cellcolor{rowgray}3.70
& \cellcolor{rowgray}\underline{0.265} & \cellcolor{rowgray}0.432 \\

& TPP-Gaze \cite{d2025tpp}
& \cellcolor{rowgray2}\underline{240} & \cellcolor{bestcell}\textbf{224}
& \cellcolor{bestcell}\textbf{2.34} & \cellcolor{bestcell}\textbf{1.68}
& \cellcolor{rowgray2}4.30 & \cellcolor{rowgray2}\underline{3.47}
& \cellcolor{rowgray2}0.225 & \cellcolor{rowgray2}\underline{0.435} \\

& Ours
& \cellcolor{bestcell}\textbf{239} & \cellcolor{rowgray}\underline{228}
& \cellcolor{rowgray}\underline{2.50} & \cellcolor{rowgray}1.95
& \cellcolor{bestcell}\textbf{4.01} & \cellcolor{bestcell}\textbf{3.40}
& \cellcolor{bestcell}\textbf{0.290} & \cellcolor{bestcell}\textbf{0.444} \\
\bottomrule
\end{tabular}
}
\label{tab:video_results_numbers}
\end{table*}

\definecolor{rowgray}{gray}{0.92}
\definecolor{rowgray2}{gray}{0.86}
\definecolor{bestcell}{gray}{0.75}

\begin{table*}[t]
\centering
\small
\caption{Additional results for 3-second predictions that do not require autoregressive adjustments for the comparison methods. Lower is better for all metrics except Maximum Temporal Correlation. Metrics are scaled as indicated in the column headers. \textbf{Bold} values indicate the best scores, while \underline{underlined} values denote the second-best scores.}
\resizebox{\textwidth}{!}{
\setlength{\tabcolsep}{12pt}
\begin{tabular}{l l cccccccc}
\toprule
\multirow{2}{*}{Dataset} & \multirow{2}{*}{Method}
& \multicolumn{2}{c}{Levenshtein $\downarrow$}
& \multicolumn{2}{c}{Disc. Fréchet ($\times 10^2$) $\downarrow$}
& \multicolumn{2}{c}{DTW ($\times 10^4$) $\downarrow$}
& \multicolumn{2}{c}{Max. Temp. Corr. $\uparrow$} \\
&
& Mean & Best
& Mean & Best
& Mean & Best
& Mean & Best \\
\midrule

\multirow{8}{*}{\textbf{DIEM}}
& DeepGaze III
& \cellcolor{rowgray}84.8 & \cellcolor{rowgray}84.8
& \cellcolor{rowgray}5.99 & \cellcolor{rowgray}3.44
& \cellcolor{rowgray}3.46 & \cellcolor{rowgray}1.58
& \cellcolor{rowgray}0.000 & \cellcolor{rowgray}0.000 \\

& DiffEye
& \cellcolor{rowgray2}81.7 & \cellcolor{rowgray2}77.7
& \cellcolor{rowgray2}9.86 & \cellcolor{rowgray2}8.36
& \cellcolor{rowgray2}6.23 & \cellcolor{rowgray2}5.11
& \cellcolor{rowgray2}0.021 & \cellcolor{rowgray2}0.325 \\

& DiffEye retrained w/ saliency
& \cellcolor{rowgray}80.9 & \cellcolor{rowgray}73.0
& \cellcolor{rowgray}4.40 & \cellcolor{rowgray}3.27
& \cellcolor{rowgray}2.22 & \cellcolor{rowgray}1.59
& \cellcolor{rowgray}\underline{0.037} & \cellcolor{rowgray}0.348 \\

& GazeFormer
& \cellcolor{rowgray2}80.4 & \cellcolor{rowgray2}74.6
& \cellcolor{rowgray2}4.54 & \cellcolor{rowgray2}3.36
& \cellcolor{rowgray2}2.36 & \cellcolor{rowgray2}1.62
& \cellcolor{rowgray2}0.012 & \cellcolor{rowgray2}0.322 \\

& HAT
& \cellcolor{rowgray}83.0 & \cellcolor{rowgray}77.9
& \cellcolor{rowgray}5.20 & \cellcolor{rowgray}3.47
& \cellcolor{rowgray}2.71 & \cellcolor{rowgray}1.97
& \cellcolor{rowgray}0.009 & \cellcolor{rowgray}0.324 \\

& Chen et al.
& \cellcolor{rowgray2}\underline{80.0} & \cellcolor{rowgray2}72.6
& \cellcolor{rowgray2}4.33 & \cellcolor{rowgray2}3.08
& \cellcolor{rowgray2}2.20 & \cellcolor{rowgray2}1.49
& \cellcolor{rowgray2}0.005 & \cellcolor{rowgray2}\underline{0.384} \\

& TPP-Gaze
& \cellcolor{rowgray}80.7 & \cellcolor{rowgray}\underline{72.0}
& \cellcolor{rowgray}\underline{3.95} & \cellcolor{rowgray}\underline{2.74}
& \cellcolor{rowgray}\underline{2.07} & \cellcolor{rowgray}\underline{1.32}
& \cellcolor{rowgray}0.029 & \cellcolor{bestcell}\textbf{0.409} \\

& Ours
& \cellcolor{bestcell}\textbf{77.8} & \cellcolor{bestcell}\textbf{67.8}
& \cellcolor{bestcell}\textbf{3.31} & \cellcolor{bestcell}\textbf{2.45}
& \cellcolor{bestcell}\textbf{1.60} & \cellcolor{bestcell}\textbf{1.21}
& \cellcolor{bestcell}\textbf{0.156} & \cellcolor{rowgray2}0.374 \\

\midrule

\multirow{8}{*}{\textbf{DHF1K}}
& DeepGaze III
& \cellcolor{rowgray}59.1 & \cellcolor{rowgray}56.7
& \cellcolor{rowgray}2.41 & \cellcolor{rowgray}1.59
& \cellcolor{rowgray}1.45 & \cellcolor{rowgray}0.941
& \cellcolor{rowgray}0.017 & \cellcolor{bestcell}\textbf{0.462} \\

& DiffEye
& \cellcolor{rowgray2}\underline{58.7} & \cellcolor{rowgray2}54.5
& \cellcolor{rowgray2}\underline{2.03} & \cellcolor{rowgray2}1.74
& \cellcolor{rowgray2}\underline{1.14} & \cellcolor{rowgray2}0.949
& \cellcolor{rowgray2}0.039 & \cellcolor{rowgray2}0.319 \\

& DiffEye retrained w/ saliency
& \cellcolor{rowgray}59.7 & \cellcolor{rowgray}55.4
& \cellcolor{rowgray}2.46 & \cellcolor{rowgray}1.79
& \cellcolor{rowgray}1.44 & \cellcolor{rowgray}1.10
& \cellcolor{rowgray}0.065 & \cellcolor{rowgray}0.354 \\

& GazeFormer
& \cellcolor{rowgray2}58.8 & \cellcolor{rowgray2}55.1
& \cellcolor{rowgray2}2.24 & \cellcolor{rowgray2}1.71
& \cellcolor{rowgray2}1.26 & \cellcolor{rowgray2}0.985
& \cellcolor{rowgray2}0.047 & \cellcolor{rowgray2}0.301 \\

& HAT
& \cellcolor{rowgray}60.3 & \cellcolor{rowgray}57.1
& \cellcolor{rowgray}3.00 & \cellcolor{rowgray}2.33
& \cellcolor{rowgray}1.50 & \cellcolor{rowgray}1.23
& \cellcolor{rowgray}0.028 & \cellcolor{rowgray}0.302 \\

& Chen et al.
& \cellcolor{rowgray2}59.0 & \cellcolor{rowgray2}54.1
& \cellcolor{rowgray2}2.34 & \cellcolor{rowgray2}1.56
& \cellcolor{rowgray2}1.37 & \cellcolor{rowgray2}0.897
& \cellcolor{rowgray2}\underline{0.075} & \cellcolor{rowgray2}0.402 \\

& TPP-Gaze
& \cellcolor{rowgray}59.2 & \cellcolor{rowgray}\underline{52.9}
& \cellcolor{rowgray}2.19 & \cellcolor{bestcell}\textbf{1.36}
& \cellcolor{rowgray}1.20 & \cellcolor{rowgray}\underline{0.770}
& \cellcolor{rowgray}0.028 & \cellcolor{rowgray}\underline{0.442} \\

& Ours
& \cellcolor{bestcell}\textbf{58.2} & \cellcolor{bestcell}\textbf{52.3}
& \cellcolor{bestcell}\textbf{1.94} & \cellcolor{rowgray2}\underline{1.36}
& \cellcolor{bestcell}\textbf{1.01} & \cellcolor{bestcell}\textbf{0.745}
& \cellcolor{bestcell}\textbf{0.109} & \cellcolor{rowgray2}0.393 \\
\bottomrule
\end{tabular}
}
\label{tab:video_results_3s}
\end{table*}

\subsection{Quantitative Comparisons}

\paragraph{\textbf{Long Window Generation}} Quantitative results for long window video gaze prediction are shown in Table~\ref{tab:video_results_numbers}.
Our method achieves the best performance in almost all evaluated metrics across both datasets.
In particular, we obtain the strongest results in Levenshtein distance, DTW, and Maximum Temporal Correlation. The low Levenshtein distance indicates that the predicted gaze path closely matches the ordering and structure of human gaze transitions. The superior DTW performance demonstrates accurate temporal alignment, even under local variations in duration or sampling density. The high Maximum Temporal Correlation further confirms that the predicted gaze trajectories follow similar temporal dynamics as human observers, capturing consistent motion patterns over time. These temporal motion patterns are particularly important for videos as content changes constantly and may only be visible for a short period of time. Together, these results indicate that the model captures both the sequential structure and the temporal dynamics of gaze behavior.

The results of the best scores show that individual samples generated by our model can closely approximate specific human gaze paths. In contrast, the consistently strong mean scores indicate that the model reliably aligns with the overall distribution and dynamics of human fixations, rather than producing occasional high-quality outliers.

Improvements in Discrete Fréchet distance are less pronounced. Discrete Fréchet emphasizes the maximum deviation along the trajectory and is dominated by local outliers. Minor timing shifts or short exploratory saccades, which are common in videos, can substantially increase the distance, even when the overall trajectory structure and temporal progression match well. As a result, Discrete Fréchet is less sensitive to global sequence similarity and temporal alignment, which are central to video gaze prediction.

\paragraph{\textbf{Three Second Window Generation}} Quantitative results show a similar trend to the full-video evaluation. Additionally, the modular baseline ``DiffEye retrained w/ saliency'' remains substantially below our method, demonstrating that the performance gain comes from the combination of saliency-aware conditioning and autoregressive diffusion rather than from saliency conditioning alone.

\subsection{Qualitative Comparisons}

\paragraph{\textbf{Visual Inspection}}
Figures~\ref{fig:gaze_full}--\ref{fig:gaze_example} show how the gaze predictions progress with the methods as the video advances. 
We further refer to the supplementary videos for a comprehensive assessment of temporal dynamics and motion characteristics.
The qualitative results support the quantitative findings. Our model produces gaze trajectories that are temporally smooth and spatially stable. The predicted paths avoid abrupt jumps and exhibit coherent motion over time. In addition, the gaze behavior responds consistently to scene content and camera motion, adapting to moving objects and global motion patterns. 
In contrast, the comparison methods often generate temporally fragmented or unstable trajectories that do not react well to content change, which explains their lower scores in sequence- and time-aware metrics.

\paragraph{\textbf{User Study}}
\begin{figure}[t]
    \centering
    \includegraphics[width=\linewidth]{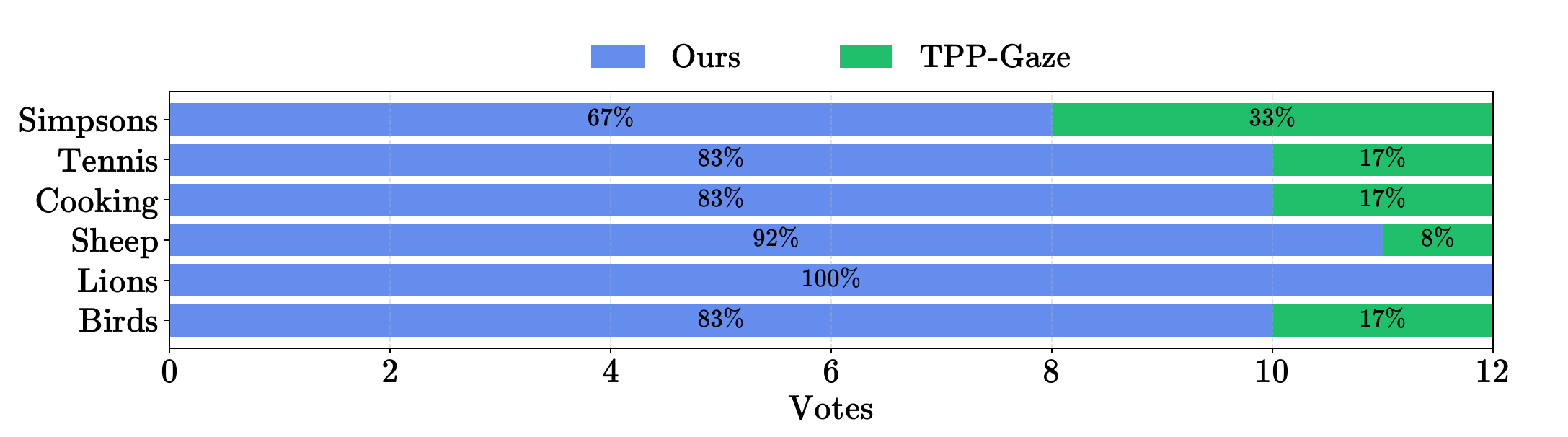}
    \caption{\textit{User study results.} Each row corresponds to one video. Our model (blue) receives consistently more votes than TPP-Gaze (green).}
    \label{fig:user_study}
\end{figure}

In the quantitative evaluation, our method achieved the best performance on all metrics except Discrete Fréchet Distance, where TPP-Gaze obtained the lowest score. To assess whether this outcome reflects limitations of our model or potential biases from the Discrete Fréchet metric for video-based gaze prediction, we conducted a qualitative user study comparing our model against TPP-Gaze.
We randomly selected six videos from the benchmark, three from each dataset. For randomly chosen ground-truth trajectories, we identified the closest predictions from both methods based on the Discrete Fréchet score. We rendered the ground truth and both model predictions as moving dot overlays on the videos (see supplementary material). The ground truth trajectory was displayed in a fixed color across all videos, while the two predicted trajectories were assigned random colors.
Twelve participants took part in the study. For each video, participants were required to watch the full sequence and could replay it before making a decision. They were asked to select the trajectory that appeared closest to the ground truth over the entire video.

The results are shown in Figure~\ref{fig:user_study}. Across all trials, our method was identified as more similar to the ground truth trajectories in 85\% of the cases. Our predictions received consistently better ratings than TPP-Gaze in every evaluated instance. These findings indicate that the inferior ranking under Discrete Fréchet Distance does not necessarily correspond to poorer perceptual quality. Instead, the metric appears less suitable for evaluating gaze prediction in videos than for static scanpath comparison tasks.

\subsection{Computational Efficiency}

Our saliency latent compression architecture was empirically chosen to preserve gaze-relevant information while reducing conditioning dimensionality and improving convergence. 
\begin{table*}[t]
\centering
\caption{Computational efficiency on DIEM validation batches using an NVIDIA RTX A6000. \textbf{Bold} values indicate the better scores.}
\label{tab:efficiency_comparison}

\resizebox{0.65\textwidth}{!}{
\setlength{\tabcolsep}{3pt}
\begin{tabular}{lccc}
\toprule
Metric & Ours & Raw Pixels & Improvement \\
\midrule
Parameters (Mil)         & \textbf{8.27}  & 32.33  & 3.91$\times$ fewer \\
Checkpoint size (MB)     & \textbf{37.89} & 348.54 & 9.20$\times$ smaller \\
\midrule
Data load time (s)       & \textbf{0.0085} & 0.6353 & 74.85$\times$ faster \\
Forward FLOPs (G)        & \textbf{1.18}   & 71.65  & 60.95$\times$ fewer \\
Throughput (samples/s)   & \textbf{49.19}  & 17.61  & 2.79$\times$ higher \\
Peak memory (MB)         & \textbf{49.50}  & 974.90 & 19.69$\times$ lower \\
\bottomrule
\end{tabular}
}

\end{table*}

\cref{tab:efficiency_comparison} compares the computational efficiency of our saliency-compressed conditioning against naive raw-pixel conditioning. 
Our compressed representation reduces the parameter count from 32.33M to 8.27M, yielding a $3.91\times$ smaller model, while decreasing checkpoint size by $9.20\times$. 
It also substantially improves runtime efficiency: data loading is $74.85\times$ faster, forward computation is reduced from 71.65G to 1.18G FLOPs, and peak memory drops from 974.90MB to 49.50MB. 
As a result, throughput increases from 17.61 to 49.19 samples/s, corresponding to a $2.79\times$ speedup.

\subsection{Ablation Studies}
\begin{table*}[t]
\centering
\small
\caption{\textit{Ablation study results}. Regular training and if not denoted otherwise is run with 90 history samples and 45 prediction samples on the DIEM dataset. Lower scores are better for all metrics except Maximum Temporal Correlation. Metrics are scaled as indicated in the column headers. \textbf{Bold} values indicate the best scores, while \underline{underlined} values denote the second-best scores.}
\resizebox{\textwidth}{!}{
\setlength{\tabcolsep}{12pt}  
\begin{tabular}{lcccccccc}
\toprule
\multirow{2}{*}{Method}
& \multicolumn{2}{c}{Levenshtein $\downarrow$}
& \multicolumn{2}{c}{Disc. Fréchet ($\times 10^2$) $\downarrow$}
& \multicolumn{2}{c}{DTW ($\times 10^4$) $\downarrow$}
& \multicolumn{2}{c}{Max. Temp. Corr. $\uparrow$} \\
& Mean & Best
& Mean & Best
& Mean & Best
& Mean & Best \\
\midrule

Ours regular
& \cellcolor{bestcell}\textbf{700.54} & \cellcolor{bestcell}\textbf{672.91}
& \cellcolor{rowgray}4.14 & \cellcolor{rowgray}3.35
& \cellcolor{bestcell}\textbf{12.33} & \cellcolor{rowgray}\underline{11.44}
& \cellcolor{bestcell}\textbf{0.235} & \cellcolor{bestcell}\textbf{0.314} \\

\midrule

10-Prediction
& \cellcolor{rowgray2}767.72 & \cellcolor{rowgray2}760.87
& \cellcolor{bestcell}\textbf{3.67} & \cellcolor{bestcell}\textbf{2.90}
& \cellcolor{rowgray2}17.31 & \cellcolor{bestcell}\textbf{6.88}
& \cellcolor{rowgray2}0.048 & \cellcolor{rowgray2}0.126 \\

90-Prediction
& \cellcolor{rowgray}727.53 & \cellcolor{rowgray}687.72
& \cellcolor{rowgray}4.38 & \cellcolor{rowgray}3.58
& \cellcolor{rowgray}14.65 & \cellcolor{rowgray}13.84
& \cellcolor{rowgray}\underline{0.230} & \cellcolor{rowgray}\underline{0.309} \\





\midrule

No initial history
& \cellcolor{rowgray2}\underline{709.15} & \cellcolor{rowgray2}\underline{681.69}
& \cellcolor{rowgray2}\underline{4.13} & \cellcolor{rowgray2}\underline{3.34}
& \cellcolor{rowgray2}\underline{12.39} & \cellcolor{rowgray2}11.57
& \cellcolor{bestcell}\textbf{0.235} & \cellcolor{bestcell}\textbf{0.314} \\

\midrule

MAGVIT2 Conditioning
& \cellcolor{rowgray2}770.99 & \cellcolor{rowgray2}770.79
& \cellcolor{rowgray2}7.97 & \cellcolor{rowgray2}6.96
& \cellcolor{rowgray2}54.92 & \cellcolor{rowgray2}48.15
& \cellcolor{rowgray2}0.054 & \cellcolor{rowgray2}0.082 \\

\bottomrule
\end{tabular}
}
\label{tab:ablations}
\end{table*}

\paragraph{\textbf{Conditioning}}
\warning{@Qi the new conditioning we tried didn't converge, is this vaild to still have as an ablation}\warning{QS: while not converging, we can still provide numbers right?}
We compare video saliency latents with a visual tokenizer-based alternative using an open-source implementation of MAGVIT2 \cite{Fang_O2-MAGVIT2}. To mirror our saliency pipeline, we converted each video frame into tokens using the pre-trained encoder. The resulting token embeddings were then applied as per-frame conditioning signals. The goal was to compress image frames into a general-purpose latent representation that could serve as conditioning without relying on saliency-specific structure. Unlike spatial saliency latents, this representation provides a discrete global summary of each frame. However, in practice, models trained with these tokenized conditioning signals failed to converge, while saliency latent conditioning trained stably. This convergence problem is also highlighted in \Cref{tab:ablations} with significantly degraded results on video gaze prediction.



\paragraph{\textbf{Prediction Length}}
We analyze the influence of the prediction horizon in the autoregressive diffusion model. The history length is fixed to 90 frames. The prediction length is varied during both training and inference.
Table~\ref{tab:ablations} summarizes the results (\textit{Prediction} rows). Short predictions (10 frames) reduce the mean Discrete Fréchet distance and achieve the best scores for several best scores. This improvement indicates that short-horizon prediction can simplify the temporal modeling problem and reduces trajectory drift. However, the mean DTW error increases substantially, which suggests reduced robustness when evaluated over full sequences.
Extending the prediction length to 90 frames degrades most metrics compared to the regular setting. The increase in Levenshtein and DTW distances indicates accumulated autoregressive error and higher temporal uncertainty over long horizons. 
Overall, we find that a 45-frames prediction provides the best trade-off between short-term accuracy and long-term stability. 


\paragraph{\textbf{Minimal Temporal Initialization}}
To evaluate the model's dependency on historical motion priors, we conduct an ablation study using a minimal initialization strategy (Table~\ref{tab:ablations}, \textit{No initial history} row). The primary experiment initially conditions the autoregressive model on a 90-frame ground-truth history. We run a round of predictions on DIEM that replaces the initial buffer with a static cold start approach by sampling only a single ground-truth coordinate and replicating it to fill the full history window. This setup effectively removes all velocity and directional cues, forcing the model to initiate predictions from a zero-motion state. 
Even under this constrained scenario, the results remain robust: both Levenshtein distance and Maximum Temporal Correlation show identical results to the baseline, suggesting that the model could recover the necessary temporal dynamics from the latent features. Only DTW decreases slightly, potentially due to an initial misalignment before the history is built through autoregression. However, the overall consistency of the results demonstrates stable and plausible gaze trajectory generation even without motion priors.



\section{Limitations and Discussions}

\paragraph{Long-Range Dependencies.} 
Modeling long-range behavioral consistency remains a challenge for gaze generation.
The proposed autoregressive diffusion model captures high-frequency temporal dynamics of gaze trajectories. However, the model relies on a sliding window of recent predictions with a fixed length. Human gaze behavior can depend on information beyond such short-term history. Long-term memory and evolving scene understanding can influence viewing behavior in dynamic scenes \cite{kruijne2016implicit}. This dependency is particularly relevant in videos with static cameras and persistent background structure \cite{yarbus2013eye}.
Future work could adopt mechanisms from video world models that maintain information over long temporal horizons \cite{wu2025video}.

\paragraph{Evaluation metrics.}
Our method achieved the highest performance in most of the quantitative analyses and human subject studies. 
One exception is the Discrete Fréchet Distance. 
This metric is typically applied to low-dimensional scanpath data and emphasizes the maximum pointwise deviation along a trajectory. Discrete Fréchet Distance is highly sensitive to local outliers. Running the metric with a dense gaze path can lead to short exploratory saccades, minor temporal shifts, or brief deviations from the ground-truth path dominating the score. It is not designed to evaluate the alignment of global temporal structure and motion trends, which is better expressed by Levenshtein distance or DTW.
The user study results further indicate that lower Discrete Fréchet scores do not necessarily correspond to better perceptual similarity.
More generally, all current trajectory metrics rely on pairwise comparisons to a single ground-truth path and do not account for the inherent stochasticity and multi-modality of human gaze. Human observers exhibit substantial inter-subject variability, and multiple qualitatively different trajectories can be equally plausible. Future work should therefore design metrics that are specifically built to quantify the perceptual validity of generated video gaze trajectories. 

\paragraph{Head movements.}
In our training datasets, the gaze trajectories were acquired with fixed head poses. While head movements may not commonly be a factor in desktop viewing settings, they may be common in physical world and wide-field displays \cite{bahill1975most, hu2020dgaze}.
Therefore, extending the current gaze trajectory towards higher-dimensional models that jointly generate head orientations and their relative gaze may shed light on application scenarios such as imitation learning and virtual/augmented reality (VR/AR).  

\section{Conclusion}
In this paper, we introduce emerging autoregressive diffusion models to generate infinite and realistic human gaze movements in videos of arbitrary length. 
We believe this work establishes a new frontier, offering a path toward generative world models that not only simulate physical environments but also understand the context-aware behavioral responses of the humans observing them.


%
\bibliographystyle{splncs04}
\bibliography{main}

@String(CVPR  = {IEEE Conf. Comput. Vis. Pattern Recog.})

@String(ICCV  = {Int. Conf. Comput. Vis.})

@String(ECCV  = {Eur. Conf. Comput. Vis.})

@String(TOG   = {ACM Trans. Graph.})

@String(CVPR  = {CVPR})

@String(ICCV  = {ICCV})

@String(ECCV  = {ECCV})

@String(TOG   = {ACM TOG})

@book{yarbus2013eye,
  title={Eye movements and vision},
  author={Yarbus, Alfred L},
  year={2013},
  publisher={Springer}
}

@article{hu2020dgaze,
  title={Dgaze: Cnn-based gaze prediction in dynamic scenes},
  author={Hu, Zhiming and Li, Sheng and Zhang, Congyi and Yi, Kangrui and Wang, Guoping and Manocha, Dinesh},
  journal={IEEE transactions on visualization and computer graphics},
  volume={26},
  number={5},
  pages={1902--1911},
  year={2020},
  publisher={IEEE}
}

@inproceedings{chen2021predicting,
  title={Predicting human scanpaths in visual question answering},
  author={Chen, Xianyu and Jiang, Ming and Zhao, Qi},
  booktitle={Proceedings of the IEEE/CVF Conference on Computer Vision and Pattern Recognition},
  pages={10876--10885},
  year={2021}
}

@inproceedings{fang2024oat,
  title={Oat: Object-level attention transformer for gaze scanpath prediction},
  author={Fang, Yini and Yu, Jingling and Zhang, Haozheng and van der Lans, Ralf and Shi, Bertram},
  booktitle={European Conference on Computer Vision},
  pages={366--382},
  year={2024},
  organization={Springer}
}

@inproceedings{nishiyasu2024gaze,
  title={Gaze scanpath transformer: Predicting visual search target by spatiotemporal semantic modeling of gaze scanpath},
  author={Nishiyasu, Takumi and Sato, Yoichi},
  booktitle={Proceedings of the IEEE/CVF Conference on Computer Vision and Pattern Recognition},
  pages={625--635},
  year={2024}
}

@inproceedings{zelinsky2019benchmarking,
  title={Benchmarking gaze prediction for categorical visual search},
  author={Zelinsky, Gregory and Yang, Zhibo and Huang, Lihan and Chen, Yupei and Ahn, Seoyoung and Wei, Zijun and Adeli, Hossein and Samaras, Dimitris and Hoai, Minh},
  booktitle={Proceedings of the IEEE/CVF Conference on Computer Vision and Pattern Recognition Workshops},
  pages={0--0},
  year={2019}
}

@inproceedings{xue2025few,
  title={Few-shot personalized scanpath prediction},
  author={Xue, Ruoyu and Xu, Jingyi and Mondal, Sounak and Le, Hieu and Zelinsky, Greg and Hoai, Minh and Samaras, Dimitris},
  booktitle={Proceedings of the IEEE/CVF Conference on Computer Vision and Pattern Recognition},
  pages={13497--13507},
  year={2025}
}

@inproceedings{yang2024unifying,
  title={Unifying top-down and bottom-up scanpath prediction using transformers},
  author={Yang, Zhibo and Mondal, Sounak and Ahn, Seoyoung and Xue, Ruoyu and Zelinsky, Gregory and Hoai, Minh and Samaras, Dimitris},
  booktitle={Proceedings of the IEEE/CVF Conference on Computer Vision and Pattern Recognition},
  pages={1683--1693},
  year={2024}
}

@article{martin2022scangan360,
  title={Scangan360: A generative model of realistic scanpaths for 360 images},
  author={Martin, Daniel and Serrano, Ana and Bergman, Alexander W and Wetzstein, Gordon and Masia, Belen},
  journal={IEEE Transactions on Visualization and Computer Graphics},
  volume={28},
  number={5},
  pages={2003--2013},
  year={2022},
  publisher={IEEE}
}

@inproceedings{d2025tpp,
  title={TPP-Gaze: modelling gaze dynamics in space and time with neural temporal point processes},
  author={D'Amelio, Alessandro and Cartella, Giuseppe and Cuculo, Vittorio and Lucchi, Manuele and Cornia, Marcella and Cucchiara, Rita and Boccignone, Giuseppe},
  booktitle={Proceedings of the Winter Conference on Applications of Computer Vision},
  pages={8775--8784},
  year={2025}
}

@article{chen2025diffusion,
  title={Diffusion forcing: Next-token prediction meets full-sequence diffusion},
  author={Chen, Boyuan and Mart{\'\i} Mons{\'o}, Diego and Du, Yilun and Simchowitz, Max and Tedrake, Russ and Sitzmann, Vincent},
  journal={Advances in Neural Information Processing Systems},
  volume={37},
  pages={24081--24125},
  year={2025}
}

@inproceedings{yin2025causvid,
    title={From Slow Bidirectional to Fast Autoregressive Video Diffusion Models},
    author={Yin, Tianwei and Zhang, Qiang and Zhang, Richard and Freeman, William T and Durand, Fredo and Shechtman, Eli and Huang, Xun},
    booktitle={CVPR},
    year={2025}
}

@inproceedings{zhang2025epona,
  author = {Zhang, Kaiwen and Tang, Zhenyu and Hu, Xiaotao and Pan, Xingang and Guo, Xiaoyang and Liu, Yuan and Huang,
  Jingwei and Yuan, Li and Zhang, Qian and Long, Xiao-Xiao and Cao, Xun and Yin, Wei},
  title = {Epona: Autoregressive Diffusion World Model for Autonomous Driving},
  booktitle = {Proceedings of the IEEE/CVF International Conference on Computer Vision (ICCV)},
  year = {2025}
}

@article{jiao25diffgaze,
	author={Jiao, Chuhan and Wang, Yao and Zhang, Guanhua and Bâce, Mihai and Hu, Zhiming and Bulling, Andreas},
	journal={ACM Transactions on Interactive Intelligent Systems}, 
	title={DiffGaze: A Diffusion Model for Modelling Fine-grained Human Gaze Behaviour on 360° Images}, 
	year={2025}}

@article{kruijne2016implicit,
  title={Implicit short-and long-term memory direct our gaze in visual search},
  author={Kruijne, Wouter and Meeter, Martijn},
  journal={Attention, Perception, \& Psychophysics},
  volume={78},
  number={3},
  pages={761--773},
  year={2016},
  publisher={Springer}
}

@inproceedings{berndt1994using,
  title={Using dynamic time warping to find patterns in time series},
  author={Berndt, Donald J and Clifford, James},
  booktitle={Proceedings of the 3rd international conference on knowledge discovery and data mining},
  pages={359--370},
  year={1994}
}

@inproceedings{aronov2006frechet,
  title={Fr{\'e}chet distance for curves, revisited},
  author={Aronov, Boris and Har-Peled, Sariel and Knauer, Christian and Wang, Yusu and Wenk, Carola},
  booktitle={European symposium on algorithms},
  pages={52--63},
  year={2006},
  organization={Springer}
}

@article{wu2025video,
  title={Video world models with long-term spatial memory},
  author={Wu, Tong and Yang, Shuai and Po, Ryan and Xu, Yinghao and Liu, Ziwei and Lin, Dahua and Wetzstein, Gordon},
  journal={arXiv preprint arXiv:2506.05284},
  year={2025}
}

@article{privitera2002algorithms,
  title={Algorithms for defining visual regions-of-interest: Comparison with eye fixations},
  author={Privitera, Claudio M. and Stark, Lawrence W.},
  journal={IEEE Transactions on pattern analysis and machine intelligence},
  volume={22},
  number={9},
  pages={970--982},
  year={2002},
  publisher={IEEE}
}

@InProceedings{Judd_2009,
  author = {Tilke Judd and Krista Ehinger and Fr{\\\'e}do Durand and Antonio Torralba},
  title = {Learning to Predict Where Humans Look},
  booktitle = {IEEE International Conference on Computer Vision (ICCV)},
  year = {2009},
  pages = {2106-2113},
  doi = {10.1109/ICCV.2009.5459462}
}

@article{borji2013state,
  title={State-of-the-art in visual attention modeling},
  author={Borji, Ali and Itti, Laurent},
  journal={IEEE transactions on pattern analysis and machine intelligence},
  volume={35},
  number={1},
  pages={185--207},
  year={2013},
  publisher={IEEE}
}

@inproceedings{bylinskii2016where,
  title={Where should saliency models look next?},
  author={Bylinskii, Zoya and Recasens, Adri{\'a} and Borji, Ali and Oliva, Aude and Torralba, Antonio},
  booktitle={European Conference on Computer Vision},
  pages={809--824},
  year={2016},
  organization={Springer}
}

@article{borji2021saliency,
  title={Saliency Prediction in the Deep Learning Era: Successes and Limitations},
  author={Borji, Ali},
  journal={IEEE Transactions on Pattern Analysis and Machine Intelligence},
  volume={43},
  number={2},
  pages={679--700},
  year={2021},
  publisher={IEEE}
}

@inproceedings{Assens2018pathgan,
  title={PathGAN: Visual Scanpath Prediction with Generative Adversarial Networks},
  author={Assens, Marc and Giro-i-Nieto, Xavier and McGuinness, Kevin and O'Connor, Noel E.},
  booktitle={ECCV Workshop on Egocentric Perception, Interaction and Computing (EPIC)},
  year={2018},
  url={https://arxiv.org/abs/1809.00567}
}

@article{Sun2019VisualSP,
  author={Wanjie Sun and Zhenzhong Chen and Feng Wu},
  journal={IEEE Transactions on Pattern Analysis and Machine Intelligence},
  title={Visual Scanpath Prediction Using IOR-ROI Recurrent Mixture Density Network},
  year={2019},
  volume={43},
  number={6},
  pages={2101-2118},
  doi={10.1109/TPAMI.2019.2957262}
}

@article{bahill1975most,
  title={Most naturally occurring human saccades have magnitudes of 15 degrees or less.},
  author={Bahill, A Terry and Adler, Deborah and Stark, Lawrence},
  journal={Investigative Ophthalmology \& Visual Science},
  volume={14},
  number={6},
  pages={468--469},
  year={1975},
  publisher={The Association for Research in Vision and Ophthalmology}
}

@article{kummerer2022deepgaze,
  title={DeepGaze III: Modeling free-viewing human scanpaths with deep learning},
  author={K{\"u}mmerer, Matthias and Bethge, Matthias and Wallis, Thomas SA},
  journal={Journal of Vision},
  volume={22},
  number={5},
  pages={7--7},
  year={2022},
  publisher={The Association for Research in Vision and Ophthalmology},
  doi={10.1167/jov.22.5.7},
  url={https://doi.org/10.1167/jov.22.5.7}
}

@inproceedings{karadiffeye,
  title={DiffEye: Diffusion-Based Continuous Eye-Tracking Data Generation Conditioned on Natural Images},
  author={Kara, Ozgur and Nisar, Harris and Rehg, James Matthew},
  booktitle={The Thirty-ninth Annual Conference on Neural Information Processing Systems},
    year={2025},
    url={https://openreview.net/forum?id=P5yoTfwyyD}
}

@inproceedings{janner2022diffuser,
  title = {Planning with Diffusion for Flexible Behavior Synthesis},
  author = {Michael Janner and Yilun Du and Joshua B. Tenenbaum and Sergey Levine},
  booktitle = {International Conference on Machine Learning},
  year = {2022},
}

@article{ho2022video,
title={Video diffusion models},
author={Ho, Jonathan and Salimans, Tim and Gritsenko, Alexey and Chan, William and Norouzi, Mohammad and Fleet, David J},
journal={arXiv:2204.03458},
year={2022}}

@article{ho2020denoising,
    title={Denoising Diffusion Probabilistic Models},
    author={Jonathan Ho and Ajay Jain and Pieter Abbeel},
    year={2020},
    journal={arXiv preprint arxiv:2006.11239}
}

@inproceedings{
  song2021scorebased,
  title={Score-Based Generative Modeling through Stochastic Differential Equations},
  author={Yang Song and Jascha Sohl-Dickstein and Diederik P Kingma and Abhishek Kumar and Stefano Ermon and Ben Poole},
  booktitle={International Conference on Learning Representations},
  year={2021},
  url={https://openreview.net/forum?id=PxTIG12RRHS}
}

@misc{rombach2021highresolution,
      title={High-Resolution Image Synthesis with Latent Diffusion Models}, 
      author={Robin Rombach and Andreas Blattmann and Dominik Lorenz and Patrick Esser and Björn Ommer},
      year={2021},
      eprint={2112.10752},
      archivePrefix={arXiv},
      primaryClass={cs.CV}
}

@article{Peebles2022DiT,
  title={Scalable Diffusion Models with Transformers},
  author={William Peebles and Saining Xie},
  year={2022},
  journal={arXiv preprint arXiv:2212.09748},
}

@misc{singer2022makeavideo,
      title={Make-A-Video: Text-to-Video Generation without Text-Video Data},
      author={Singer, Ofir and Polyak, Gal and Hayes, Thomas and Ping, Xi and Chen, Shang-Fu and Ding, Ying and Yuan, Sheng and Adejuwon, Elijah and Korthy, Justin and Tesfaye, Hadar and others},
      year={2022},
      eprint={2209.14792},
      archivePrefix={arXiv},
      primaryClass={cs.CV},
      url={https://arxiv.org/abs/2209.14792}
}

@misc{openai2024sora,
  author       = {{OpenAI}},
  title        = {Video generation models as world simulators},
  howpublished = {\url{https://openai.com/index/video-generation-models-as-world-simulators/}},
  year         = {2024},
  note         = {Accessed: 3/3/2026}
}

@inproceedings{dong2024diffuserlite,
  title={DiffuserLite: Towards Real-time Diffusion Planning},
  author={Dong, Zibin and Hao, Jianye and Li, Pengyi and Ni, Fei and Wang, Yitian and Yuan, Yifu and Zheng, Yan},
  booktitle={Advances in Neural Information Processing Systems},
  volume={37},
  pages={122556--122585},
  year={2024},
  url={https://proceedings.neurips.cc/paper_files/paper/2024/hash/dd6a47bc0aad6f34aa5e77706d90cdc4-Abstract-Conference.html}
}

@InProceedings{gazeformer,
    author    = {Mondal, Sounak and Yang, Zhibo and Ahn, Seoyoung and Samaras, Dimitris and Zelinsky, Gregory and Hoai, Minh},
    title     = {Gazeformer: Scalable, Effective and Fast Prediction of Goal-Directed Human Attention},
    booktitle = {Proceedings of the IEEE/CVF Conference on Computer Vision and Pattern Recognition (CVPR)},
    month     = {June},
    year      = {2023},
    pages     = {1441-1450}
}

@article{deepgaze3,
    author = {Kümmerer, Matthias and Bethge, Matthias and Wallis, Thomas S. A.},
    title = {DeepGaze III: Modeling free-viewing human scanpaths with deep learning},
    journal = {Journal of Vision},
    volume = {22},
    number = {5},
    pages = {7-7},
    year = {2022},
    month = {04},
    issn = {1534-7362},
    doi = {10.1167/jov.22.5.7},
    url = {https://doi.org/10.1167/jov.22.5.7},
    eprint = {https://arvojournals.org/arvo/content_public/journal/jov/938587/i1534-7362-22-5-7_1650885429.74577.pdf},
}

@inproceedings{mit1003,
  title={Learning to predict where humans look},
  author={Judd, Tilke and Ehinger, Krista and Durand, Fr{\'e}do and Torralba, Antonio},
  booktitle={2009 IEEE 12th international conference on computer vision},
  pages={2106--2113},
  year={2009},
  organization={IEEE}
}

@InProceedings{UNISAL,
author="Droste, Richard
and Jiao, Jianbo
and Noble, J. Alison",
editor="Vedaldi, Andrea
and Bischof, Horst
and Brox, Thomas
and Frahm, Jan-Michael",
title="Unified Image and Video Saliency Modeling",
booktitle="Computer Vision -- ECCV 2020",
year="2020",
publisher="Springer International Publishing",
address="Cham",
pages="419--435"
}

@inproceedings{gaze360dataset,
author = {Xu, Yanyu and Dong, Yanbing and Wu, Junru and Sun, Zhengzhong and Shi, Zhiru and Yu, Jingyi and Gao, Shenghua},
year = {2018},
month = {06},
pages = {5333-5342},
title = {Gaze Prediction in Dynamic 360° Immersive Videos},
doi = {10.1109/CVPR.2018.00559}
}

@article{salientstreamnguyen,
  title={Enhancing 360 video streaming through salient content in head-mounted displays},
  author={Nguyen, Anh and Yan, Zhisheng},
  journal={Sensors},
  volume={23},
  number={8},
  pages={4016},
  year={2023},
  publisher={MDPI}
}

@inproceedings{attention360video,
  title={Your attention is unique: Detecting 360-degree video saliency in head-mounted display for head movement prediction},
  author={Nguyen, Anh and Yan, Zhisheng and Nahrstedt, Klara},
  booktitle={Proceedings of the 26th ACM international conference on Multimedia},
  pages={1190--1198},
  year={2018}
}

@inproceedings{scanddm,
title = "ScanDMM: A Deep Markov Model of Scanpath Prediction for 360◦ Images",
author = "Xiangjie Sui and Yuming Fang and Hanwei Zhu and Shiqi Wang and Zhou Wang",
note = "Full text of this publication does not contain sufficient affiliation information. With consent from the author(s) concerned, the Research Unit(s) information for this record is based on the existing academic department affiliation of the author(s).; 2023 IEEE/CVF Conference on Computer Vision and Pattern Recognition (CVPR 2023), CVPR2023 ; Conference date: 18-06-2023 Through 22-06-2023",
year = "2023",
doi = "10.1109/CVPR52729.2023.00675",
language = "English",
isbn = "979-8-3503-0130-4",
series = "Proceedings of the IEEE Computer Society Conference on Computer Vision and Pattern Recognition",
publisher = "IEEE",
pages = "6989--6999",
booktitle = "Proceedings - 2023 IEEE/CVF Conference on Computer Vision and Pattern Recognition, CVPR 2023",
address = "United States",
url = "https://cvpr2023.thecvf.com/Conferences/2023, https://openaccess.thecvf.com/menu, https://ieeexplore.ieee.org/xpl/conhome/1000147/all-proceedings",
}

@ARTICLE{scangan360,
  author={Martin, Daniel and Serrano, Ana and Bergman, Alexander W. and Wetzstein, Gordon and Masia, Belen},
  journal={IEEE Transactions on Visualization and Computer Graphics}, 
  title={ScanGAN360: A Generative Model of Realistic Scanpaths for 360° Images}, 
  year={2022},
  volume={28},
  number={5},
  pages={2003-2013},
  doi={10.1109/TVCG.2022.3150502}}

@misc{scanpathcodelength,
      title={Scanpath Prediction in Panoramic Videos via Expected Code Length Minimization}, 
      author={Mu Li and Kanglong Fan and Kede Ma},
      year={2025},
      eprint={2305.02536},
      archivePrefix={arXiv},
      primaryClass={cs.CV},
      url={https://arxiv.org/abs/2305.02536}, 
}

@inproceedings{wang2018revisiting,
  title={Revisiting video saliency: A large-scale benchmark and a new model},
  author={Wang, Wenguan and Shen, Jianbing and Guo, Fang and Cheng, Ming-Ming and Borji, Ali},
  booktitle={Proceedings of the IEEE Conference on computer vision and pattern recognition},
  pages={4894--4903},
  year={2018}
}

@article{mital2011clustering,
  title={Clustering of gaze during dynamic scene viewing is predicted by motion},
  author={Mital, Parag K and Smith, Tim J and Hill, Robin L and Henderson, John M},
  journal={Cognitive computation},
  volume={3},
  number={1},
  pages={5--24},
  year={2011},
  publisher={Springer}
}

@article{song2020denoising,
  title={Denoising diffusion implicit models},
  author={Song, Jiaming and Meng, Chenlin and Ermon, Stefano},
  journal={arXiv preprint arXiv:2010.02502},
  year={2020}
}

@inproceedings{scantd,
author = {Wang, Yujia and Zhang, Fang-Lue and Dodgson, Neil A.},
title = {ScanTD: 360° Scanpath Prediction based on Time-Series Diffusion},
year = {2024},
isbn = {9798400706868},
publisher = {Association for Computing Machinery},
address = {New York, NY, USA},
url = {https://doi.org/10.1145/3664647.3681315},
doi = {10.1145/3664647.3681315},
booktitle = {Proceedings of the 32nd ACM International Conference on Multimedia},
pages = {7764–7773},
numpages = {10},
location = {Melbourne VIC, Australia},
series = {MM '24}
}

@inproceedings{VTP360video,
author = {Chao, Fang-Yi and Ozcinar, Cagri and Smolic, Aljosa},
year = {2021},
month = {10},
pages = {},
title = {Transformer-based Long-Term Viewport Prediction in 360° Video: Scanpath is All You Need},
doi = {10.1109/MMSP53017.2021.9733647}
}

@article{scanpathmultimodalfusion,
title = {Scanpath prediction in panoramic videos through multimodal fusion},
journal = {Displays},
volume = {91},
pages = {103199},
year = {2026},
issn = {0141-9382},
doi = {https://doi.org/10.1016/j.displa.2025.103199},
url = {https://www.sciencedirect.com/science/article/pii/S0141938225002367},
author = {Yucheng Zhu and Yu Wang and Weimin Zhang and Jialiang Chen and Yunhao Li}
}

@misc{Fang_O2-MAGVIT2,
author = {Fang, Xuezhi and Yao, Yiqun and Jiang, Xin and Li, Xiang and Yu, Naitong and Wang, Yequan},
license = {Apache-2.0},
title = {O2-MAGVIT2},
year = {2024},
url = {https://github.com/cofe-ai/O2-MAGVIT2}
}

@inproceedings{jiang2018deepvs,
  title={Deepvs: A deep learning based video saliency prediction approach},
  author={Jiang, Lai and Xu, Mai and Liu, Tie and Qiao, Minglang and Wang, Zulin},
  booktitle={Proceedings of the european conference on computer vision (eccv)},
  pages={602--617},
  year={2018}
}

@article{wang2019revisiting,
  title={Revisiting video saliency prediction in the deep learning era},
  author={Wang, Wenguan and Shen, Jianbing and Xie, Jianwen and Cheng, Ming-Ming and Ling, Haibin and Borji, Ali},
  journal={IEEE transactions on pattern analysis and machine intelligence},
  volume={43},
  number={1},
  pages={220--237},
  year={2019},
  publisher={IEEE}
}

@inproceedings{zhang2023human,
  title={The human gaze helps robots run bravely and efficiently in crowds},
  author={Zhang, Qianyi and Hu, Zhengxi and Song, Yinuo and Pei, Jiayi and Liu, Jingtai},
  booktitle={2023 IEEE international conference on robotics and automation (ICRA)},
  pages={7540--7546},
  year={2023},
  organization={IEEE}
}

@article{li2022fusion,
  title={Fusion of human gaze and machine vision for predicting intended locomotion mode},
  author={Li, Minhan and Zhong, Boxuan and Lobaton, Edgar and Huang, He},
  journal={IEEE Transactions on Neural Systems and Rehabilitation Engineering},
  volume={30},
  pages={1103--1112},
  year={2022},
  publisher={IEEE}
}

@article{patney2016towards,
  title={Towards foveated rendering for gaze-tracked virtual reality},
  author={Patney, Anjul and Salvi, Marco and Kim, Joohwan and Kaplanyan, Anton and Wyman, Chris and Benty, Nir and Luebke, David and Lefohn, Aaron},
  journal={ACM Transactions On Graphics (TOG)},
  volume={35},
  number={6},
  pages={1--12},
  year={2016},
  publisher={ACM New York, NY, USA}
}

@inproceedings{po2025long,
  title={Long-context state-space video world models},
  author={Po, Ryan and Nitzan, Yotam and Zhang, Richard and Chen, Berlin and Dao, Tri and Shechtman, Eli and Wetzstein, Gordon and Huang, Xun},
  booktitle={Proceedings of the IEEE/CVF International Conference on Computer Vision},
  pages={8733--8744},
  year={2025}
}

@inproceedings{zeng2025foveated,
  title={Foveated instance segmentation},
  author={Zeng, Hongyi and Liu, Wenxuan and Xia, Tianhua and Chen, Jinhui and Li, Ziyun and Zhang, Sai Qian},
  booktitle={Proceedings of the Computer Vision and Pattern Recognition Conference},
  pages={24496--24505},
  year={2025}
}

@article{ding2025understanding,
  title={Understanding world or predicting future? a comprehensive survey of world models},
  author={Ding, Jingtao and Zhang, Yunke and Shang, Yu and Zhang, Yuheng and Zong, Zefang and Feng, Jie and Yuan, Yuan and Su, Hongyuan and Li, Nian and Sukiennik, Nicholas and others},
  journal={ACM Computing Surveys},
  volume={58},
  number={3},
  pages={1--38},
  year={2025},
  publisher={ACM New York, NY}
}

@article{ricker2010loss,
  title={Loss of visual working memory within seconds: the combined use of refreshable and non-refreshable features.},
  author={Ricker, Timothy J and Cowan, Nelson},
  journal={Journal of experimental psychology: Learning, memory, and cognition},
  volume={36},
  number={6},
  pages={1355},
  year={2010},
  publisher={American Psychological Association}
}
\end{document}